%% file: main.tex
\documentclass{article} 
\usepackage{iclr2025_conference,times}

\input{math_commands.tex}

\usepackage{hyperref}
\usepackage{url}
\usepackage{graphicx}

\title{Finding Stable Subnetworks at Initialization with Dataset Distillation}

\author{Luke McDermott \\
UC San Diego\\
\texttt{lmcdermo@ucsd.edu} \\
\And
Rahul Parhi \\
UC San Diego \\
\texttt{rahul@ucsd.edu}
}

\iclrfinalcopy 
\begin{document}

\maketitle

\begin{abstract}
Recent works have shown that Dataset Distillation, the process for summarizing the training data, can be leveraged to accelerate the training of deep learning models. However, its impact on training dynamics, particularly in neural network pruning, remains largely unexplored. In our work, we use distilled data in the inner loop of iterative magnitude pruning to produce sparse, trainable subnetworks at initialization --- more commonly known as \textit{lottery tickets}. While using 150x less training points, our algorithm matches the performance of traditional lottery ticket rewinding on ResNet-18 \& CIFAR-10. Previous work highlights that lottery tickets can be found when the dense initialization is stable to SGD noise (i.e. training across different ordering of the data converges to the same minima). We extend this discovery, demonstrating that stable subnetworks can exist even within an unstable dense initialization. In our linear mode connectivity studies, we find that pruning with distilled data discards parameters that contribute to the sharpness of the loss landscape. Lastly, we show that by first generating a stable sparsity mask at initialization, we can find lottery tickets at significantly higher sparsities than traditional iterative magnitude pruning.
\end{abstract}

\section{Introduction}
Sparse neural networks play a crucial role in deep learning research, enhancing model inference efficiency to reduce costs and environmental impacts. This is especially true as foundational models continue to grow in parameter count. However, many iterative pruning \citep{PruneSurvey} methods require extensive retraining, which is computationally prohibitive for larger models. In parallel, researchers have been exploring how synthetic data representations such as those generated by dataset distillation methods \citep{DatasetDistillationSurvey} can be leveraged to reduce the training set size, accelerating the training process. Motivated by this, we investigate the training dynamics of sparse initializations generated by dataset distillation. We aim to learn how smaller synthetic datasets can be used as a proxy for real data, allowing us to efficiently prune at intialization.

In this work, we employ Iterative Magnitude Pruning (IMP) with weight rewinding which trains a model to convergence, prunes the weights with the lowest magnitude after training, then rewinds the non-pruned weights back to their original values at initialization. The next iteration continues the process: training the --now sparse-- initialization, pruning the weights, and rewinding until we achieve the desired sparsity. The final output of this algorithm is a sparsity mask of ones and zeros, depicting which weights are to be pruned. We can apply this mask to the original initialization and sparsely train the model. This iterative process "cheats" by leveraging post-training information to decide which weights are unimportant and can be pruned. However, the method breaks when the sparsely retrained model lands in a different minima, as the sparsity mask is uninformed about this new area of the loss landscape. 

To study if sparse training can even achieve equivalent performance of the dense model, \citet{LTH} leverage IMP with weight rewinding as a brute force approach to finding such sparse initializations. In this work, they propose the Lottery Ticket Hypothesis \cite{LTH} which states that under some conditions, a dense, randomly initialized model contains a subnetwork that can be trained in isolation to achieve the same performance as the dense model. This subnetwork is denoted as a \textit{lottery ticket}. IMP cannot always find lottery tickets; \citet{LMCLTH} show that lottery tickets exist when the dense model is stable to SGD noise. That is, the dense model can be trained across different shufflings of the data and converge to the same linearly connected mode during SGD. This only occurs on smaller datasets and models. \citet{LMCLTH} propose that instead of rewinding the model weights back to intialization, they must rewind to an early point in training, when the dense model is stable. These subnetworks are no longer \textit{lottery tickets}, instead referred to as \textit{matching subnetworks}, as the sparsity mask must be applied to the model initialization to be considered sparse training from initialization.

Rather than spawning in at a stable point in training \citep{spawn}, we have found that stable, sparse subnetworks can exist in an unstable dense initialization. By summarizing the training data into only 1-50 synthetic images per class with dataset distillation, we can perform the iterative retraining over the small, yet powerful synthetic dataset. The sparsity masks chosen by distilled data observe greater stability than the dense model or any traditionally generated sparsity masks. We refer to subnetworks chosen by distilled data as \textit{synthetic subnetworks}. 

In this work, we conduct a large instability analysis through linear mode connectivity studies, loss landscape visualizations, and a Hessian analysis. We show that synthetic subnetworks can match performance with traditional IMP on ResNet-18 and CIFAR-10, even when using 150x less training points. Lastly, we combine these methods: leveraging distilled pruning to find a stable sparse initialization that traditional IMP can use. We observe lottery tickets with 10x fewer parameters than previously obtainable on subsets of ImageNet.

\section{Related Work}

\subsection{Sparsity in Neural Networks}
Neural Network Pruning \citep{PruneSurvey, Sparseland} aims to remove weights in a model to reduce computational costs. Popular methods primarily target inference efficiency, often leveraging a larger overhead to search for unimportant weights before deploying the model. However, pruning before training \citep{SNIP}, during training, or after training \citep{HanPrune} has been well studied. In order to reduce the cost of training as well, \citet{SNIP, SynFlow} explore how to prune at initialization. This is essentially producing optimal sparse architectures for downstream tasks, the end goal for almost any pruning research. Despite these great ambitions, pruning at initialization does not perform as we hope \citep{PruneInitWhyMissing}. 

\citet{LTH} have empirically shown through brute force that a high performing sparse subnetwork can exist with the Lottery Ticket Hypothesis; however, finding such a subnetwork efficiently at initialization is an open problem. Furthermore, this empirical proof of existence only holds on small models and datsets. \citet{LMCLTH} find that this hypothesis only holds when the dense model is "stable" to the effects of data shuffling. At initialization, only the smallest computer vision models are stable to SGD noise without weight permutations \citep{gitrebasin, permutation1}. Again, a model is deemed stable if training across different orders of training data result in models with linear mode connectivity. 

Previous work has shown that trained dense models are connected in the loss landscape through nonlinear paths \citep{NonlinearPath1, NonlinearPath2, NonlinearPath3}. However, linear paths or Linear Mode Connectivity (LMC) are a relatively uncommon phenomenon for two models that train from the same initialization. Without strong intervention like permuting the trained weights \citep{gitrebasin}, this occurs in rare cases, such as training small models like MLPs on subsets of MNIST. Otherwise, a model can observe linear mode connectivity after some initial training. In this work, we aim explore if subnetworks of dense models can show improved stability at initialization, rather than early in training like more modern lottery ticket work \citep{DataDiet}.

\subsection{Dataset Distillation}
We explore the idea of synthetic data in pruning by using dataset distillation methods. In general, dataset distillation optimizes a small, synthetic dataset to match the performance of a model trained on real data. This bi-level optimization problem can be defined as minimizing the difference of average loss over all validation points:

\begin{equation}
    \argmin_{\mathcal{D}_{\text{syn}}} | L(\Phi(\mathcal{D}_{\text{real}}); \mathcal{D}_{\text{val}}) - L(\Phi(\mathcal{D}_{\text{syn}}); \mathcal{D}_{\text{val}}) | 
\end{equation}

where $\Phi$ is a training algorithm returning optimal parameters for the dataset $\mathcal{D}_{\text{syn}}$. Many methods include matching the training trajectories of a student and teacher model \citep{MTT}, meta model matching \citep{MetaModel}, or factoring \citep{Factoring}. In this work, we focus on distillation methods that match the training trajectory as we primarily are interested in approximating the converged minima of SGD on the real data. For a simplified view of the distillation process on classification tasks, these methods compress each class of images down to a few samples, removing image-specific noise. The synthetic dataset maximizes the amount of label information in the constrained set, similar to the Information Bottleneck \cite{IB,DeepIB}.

Current dataset distillation methods do not properly maintain relevant information on the downstream task for harder datasets, but they do sufficiently minimize the image-specific information resulting in over-compression. These issues arise as a result of the high computational complexity of optimizing a synthetic training set. This limits current methods to smaller datasets CIFAR and MNIST. For an example, we use Efficient Dataset Condensation \citep{EfficientDatasetCondensation} in this paper which achieves 50.6-74.5\% accuracy on CIFAR-10 depending the the size of the synthetic dataset (1,10, or 50 images per class). Recently \citet{imagenetdistillation} extend this to ImageNet; however, the performance is still far away from using this to replace the real training set. In addition to lack of support for larger datasets, distilled data does not generalize well across different neural architectures as training trajectories only match in architectures the data was optimized for. Despite these challenges, this is a new field that has made immense progress over the past few years. We believe this work should help showcase potential applications of such data. 

While the aim of dataset distillation literature is to have a small synthetic set to improve training efficiency on other architectures, less aggressive compression may be needed to maintain performance. In this paper, we show that training efficiency is not the only use case for distilled data. This data can be used for architectural validation like Neural Architecture Search \cite{nassurvey, morph} or pruning where high training performance is less essential.

\section{Dataset Distillation for Neural Network Pruning.} 
We employ Iterative Magnitude Pruning with Dataset Distillation for the first time to better understand how synthetic data can be used as a proxy for estimating the importance of each parameter. Specifically, we use dataset distillation methods that match training trajectories to ensure that training on synthetic data yields similar converged results to training on real data \citep{EfficientDatasetCondensation}. We refer to this algorithm as \textit{Distilled Pruning}. 

\begin{figure}[htb]
  \centering
  \includegraphics[width=3.25in]{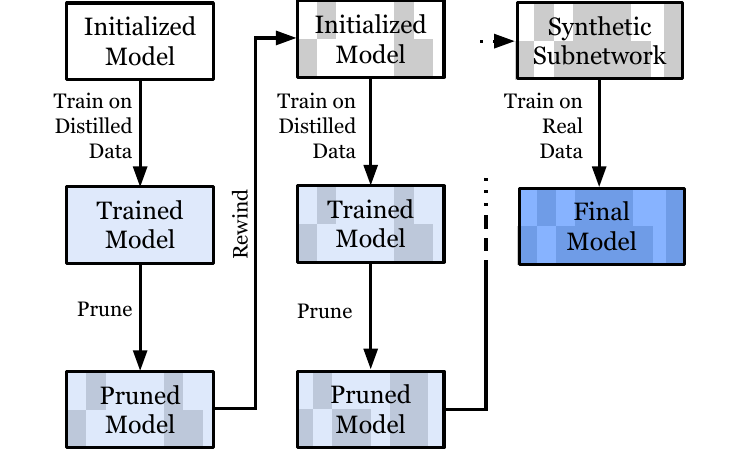}
  \caption{Distilled Pruning Algorithm Diagram}
  \label{fig:distilledpruning}
\end{figure}

\begin{figure}[htb]
\centering
  \includegraphics[width=12cm]{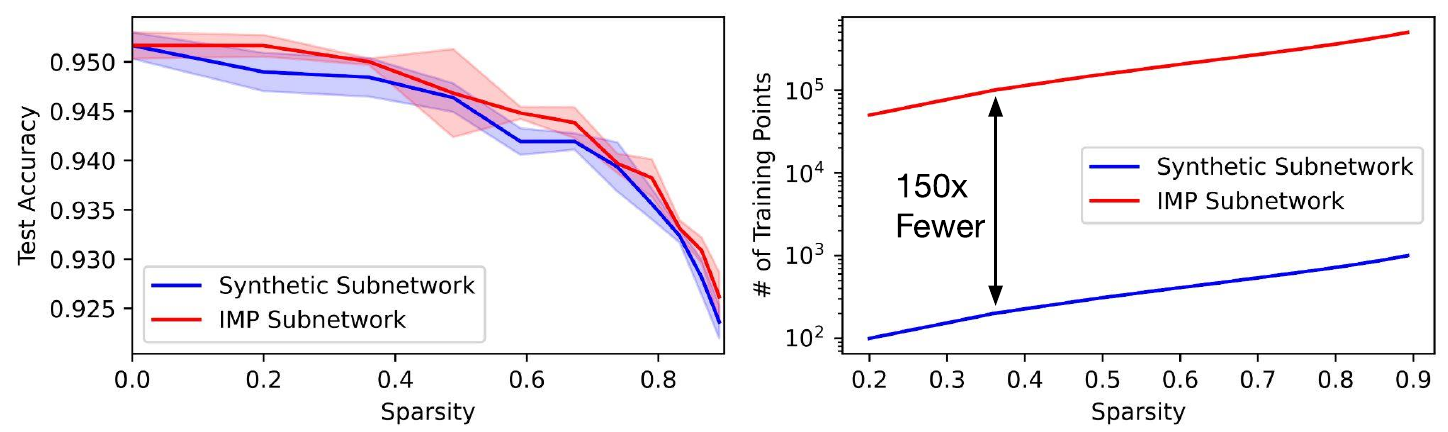}
  \caption{Performance of Distilled Pruning vs Traditional IMP on ResNet-18 \& CIFAR-10. The distilled dataset consisted of 10 images per class. Error bars are plotted as we average across 4 seeds. The plot on the right measures the amount of data points used in training to find a sparsity mask at x sparsity. Note that in IMP we are not matching the dense performance since we rewind back to initialization for both methods --- not to an early point in training. Lottery tickets do not exist here past $\approx$ 40\% sparsity.}
  \label{fig:SVATTM}
\end{figure}

As shown in Figure \ref{fig:distilledpruning}, to obtain a suitable sparsity mask, we train the network to convergence on distilled data, prune the lowest 20\% of weights by magnitude, rewind the remaining weights to their initialization, and repeat until the desired sparsity is reached. The final model should have the same randomly initialized weights, except with a sparsity mask chosen by the synthetic data. We refer to subnetworks found with distilled data as \textit{synthetic subnetworks} and those with real data as \textit{IMP subnetworks}. We can train synthetic subnetworks on real data to achieve sufficient performance at high sparsities, just like IMP subnetworks. Figure \ref{fig:SVATTM} showcases the performance of this method compared to traditional IMP on CIFAR-10 with ResNet-18.

\section{Stability of Subnetworks.}
\begin{figure}[htb]
  \centering
  \includegraphics[width=5.5in]{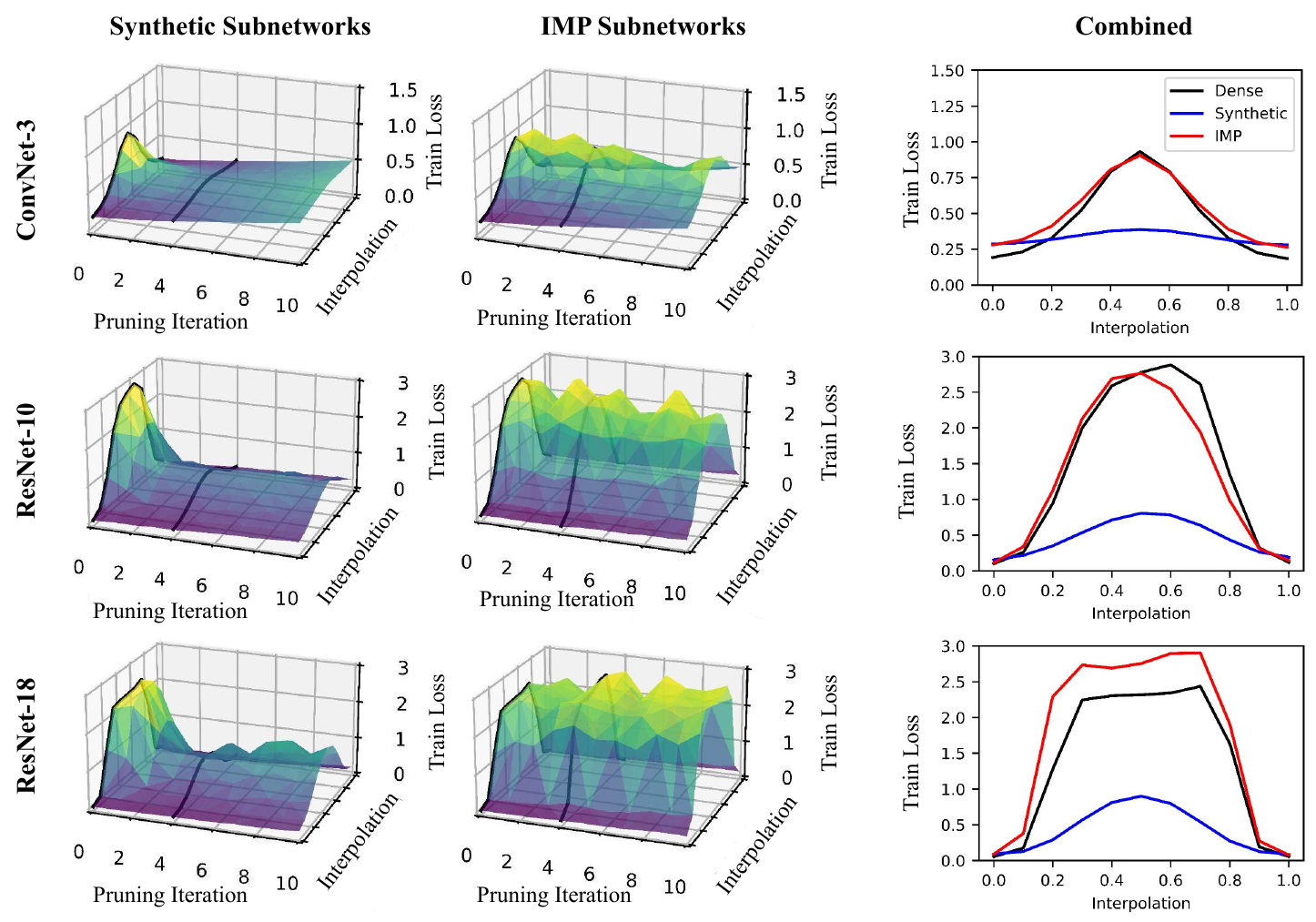}
  \caption{Comparison of the stability of synthetic vs. IMP subnetworks at initialization on CIFAR-10. We show how the loss increases as you interpolate the weights between two trained models. We measure this for subnetworks of different sparsities. The left column is reserved for subnetworks found via distilled data, and the middle column is for subnetworks found with real data. We aggregate all the information in each row for a better comparison. The dark lines in the 3D plots represents the pruning iteration we used for the combined plot; the dense model is iteration 0. }
  \label{fig:LMC_cifar10}
\end{figure}

\begin{figure}[htb]
  \centering
  \makebox[0pt]{\includegraphics[width=14cm]{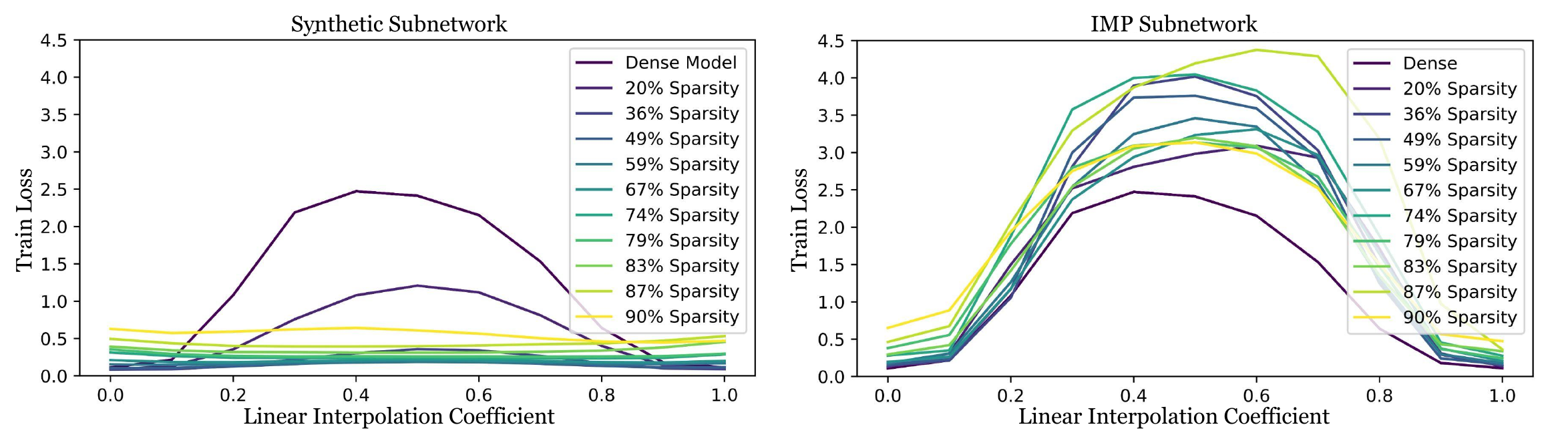}}
  \caption{Comparison of the stability of synthetic vs. IMP subnetworks at initialization on ImageNet-10 and ResNet-10. An increased loss across interpolation implies instability / trained networks landing in different minima. }
  \label{fig:LMC_imagenet10}
\end{figure}

To understand the training dynamics of these sparse subnetworks at intialization, we conduct an instability analysis similar to \citet{LMCLTH}. We take a randomly initialized model, generate a sparsity mask with a desired pruning method, and train the subnetwork across two different orderings of the real training data. We interpolate between the weights of the two model variants, measuring the training loss at each point in the interpolation as shown in Figures \ref{fig:LMC_cifar10} and \ref{fig:LMC_imagenet10}. We assess the linear mode connectivity of these subnetworks, determining if the model is stable to SGD noise. If the loss does not increase during interpolation, then this implies the two trained versions exist in the same minima or at least the same flat basin. Otherwise, if the loss increases, then there exists a barrier and the two trained models found different minima. When the models do not converge to the same minima, post-training information of the model weights are not as useful since the next retraining stage will land in a different area of the loss landscape. 

In Figures \ref{fig:LMC_cifar10} and \ref{fig:LMC_imagenet10}, we observe full linear mode connectivity in simpler scenarios such as ConvNet-3 on CIFAR-10 and ResNet-10 on ImageNet-10. In these cases, the unstable dense model contains a stable, sparse subnetwork at initialization. More importantly, traditional IMP is not able to produce stable subnetworks in these settings. As we iteratively prune over distilled data, the results exhibit more stability despite lower trainability, as shown with higher, yet flatter, losses. We postulate that the parameters pruned on distilled data, yet still exist in the IMP subnetwork, capture the outlier information of the real data which contribute to a sharper, but more trainable, landscape. Since IMP subnetworks are not stable, the learned outlier information is dependent on the order during training.

Despite the glaring differences in stability between the two classes of subnetworks, we could not find a noticeable structure or pattern in the sparsity masks. IMP and synthetic subnetworks from the same initialization are nearly independent when judging the distribution of zeros in these masks. We also observed that the magnitudes of pruned weights at initialization share the same distribution between methods.

\subsection{Loss Landscape Analysis}
\begin{figure}[htb]
\centering
  \includegraphics[width=12cm]{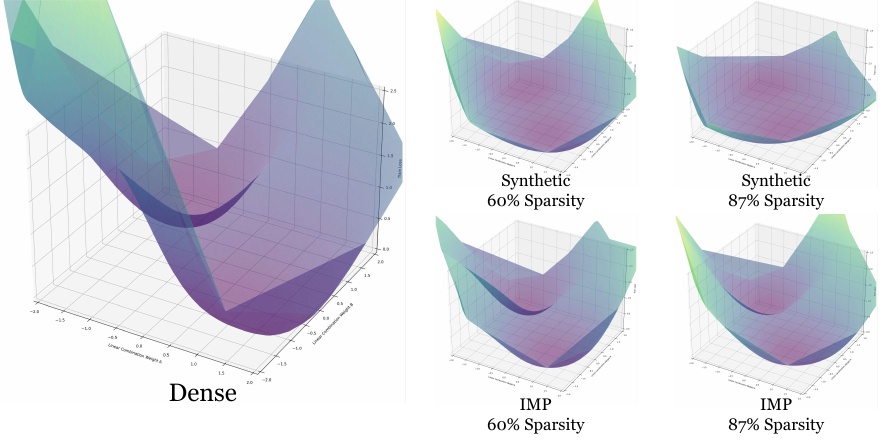}
  \caption{Loss Landscape visualization around the neigbhorhood defined by trained models on different seeds for ConvNet-3 and CIFAR-10.}
  \label{fig:loss_landscape}
\end{figure}

While linear mode connectivity is useful to study the loss landscape, the lightweight method can only show us a one-dimensional slice of the bigger picture. We further examine the landscapes across two dimensions of parameters as shown in Figure \ref{fig:loss_landscape}. 

In this study, we train the same network across 2 different seeds, much like in linear interpolation; however, we also created 2 orthogonal vectors from this to sample nearby points. This maps the hyperdimensional parameter space down to two dimensions, since every point shown in our visualization is a linear combination of the two vectors. For each plot, we sampled 10,000 points in the 2-D space and evaluated the loss on real training data. 


From these visualizations, IMP chooses subnetworks that exhibit a similar landscape to the dense model. We see the trained models fall into two separate minima in both the IMP and Dense cases, explaining the loss barrier in the Figure \ref{fig:LMC_cifar10}. Subnetworks chosen with distilled data are falling into the same flat basin. 

\subsection{Hessian Analysis}
For a more quantitative approach to measuring sharpness of the loss landscape, we measure the diagonal of the Hessian over batches of real training data --- as computing the full Hessian is computationally prohibitive. We study the diagonal specifically for ConvNet-3 on CIFAR-10 due to the compute costs scaling with the number of parameters.

\begin{center}
\begin{tabular}{||c | c | c | c | c ||} 
 \hline
 Subnetwork & Sparsity & Min/Max & Mean $\pm$ Std & Avg Magnitude \\ [0.5ex] 
 \hline\hline
 Dense & 0 & -1.847 / 1.344 & $.0048 \pm .035$ & .0067\\ 
 \hline
 IMP & 60\% & -3.073  / 4.510  & $.0051 \pm .0559$ & .0087\\
 \hline
 IMP & 90\% & -6.323 / 6.127  & $.0035 \pm .0594$ & .0060\\
 \hline
 Synthetic & 60\% & -0.982 / 2.093 & $.0047 \pm .0286$ & .0061 \\
 \hline
 Synthetic & 90\% & -2.338 / 1.730  & $.0037 \pm .0354$ & .0053\\ [1ex] 
 \hline
\end{tabular}
\end{center}

As shown in the table, synthetic subnetworks have a tighter distribution of diagonal hessian values around zero than both the dense model and IMP subnetworks.  We see that across every measure, small amounts of pruning with traditional IMP leads to a sharper landscape over the dense model. With our synthetic subnetworks, the distribution is much closer to zero with less variance. Across both methods, higher sparsities generally improve the smoothness; however, there is still a noticeable difference between synthetic and IMP-chosen subnetworks.


\section{Overview of Distilled Pruning vs. IMP}
The compressed nature of distilled data plays a critical role in stability. We examine how varying the compression rate affects both the performance of synthetic subnetworks and their stability. Across each dataset, we compare the performance of distilled pruning to traditional IMP by measuring the ratio of the synthetic subnetwork accuracy over the IMP accuracy. Performance ratios greater than one imply that synthetic subnetworks have higher accuracies while ratios less than one imply they underperform.

Similarly, we compare the stability by measuring the ratio of the loss barrier height of synthetic and IMP subnetworks. The barrier height is measured as the distance between the trained loss and the loss of the interpolated model from our linear mode connectivity experiments. A low barrier height means the trained variants are linearly mode connected.

Both a low (60\%) and high (87\%) sparsity example is reported for each setting. Since each point measures the stability and performance of a synthetic subnetwork \textit{compared to the IMP subnetwork}, we place a single point for IMP that represents a performance and stability ratio of 1.

We report various compression ratios, models, and classes. The data compression is calculated by the original dataset size divided by the distilled dataset size. For example, CIFAR-10 contains 5000 training points per class that can be distilled to 10 images per class (ipc) resulting in a compression ratio of 500. For CIFAR-100 only contains 500 training points per class, so distilling this to 10 ipc is only a 50x compression ratio.

\begin{figure}[htb]
  \centering
  \makebox[0pt]{\includegraphics[width=5.5in]{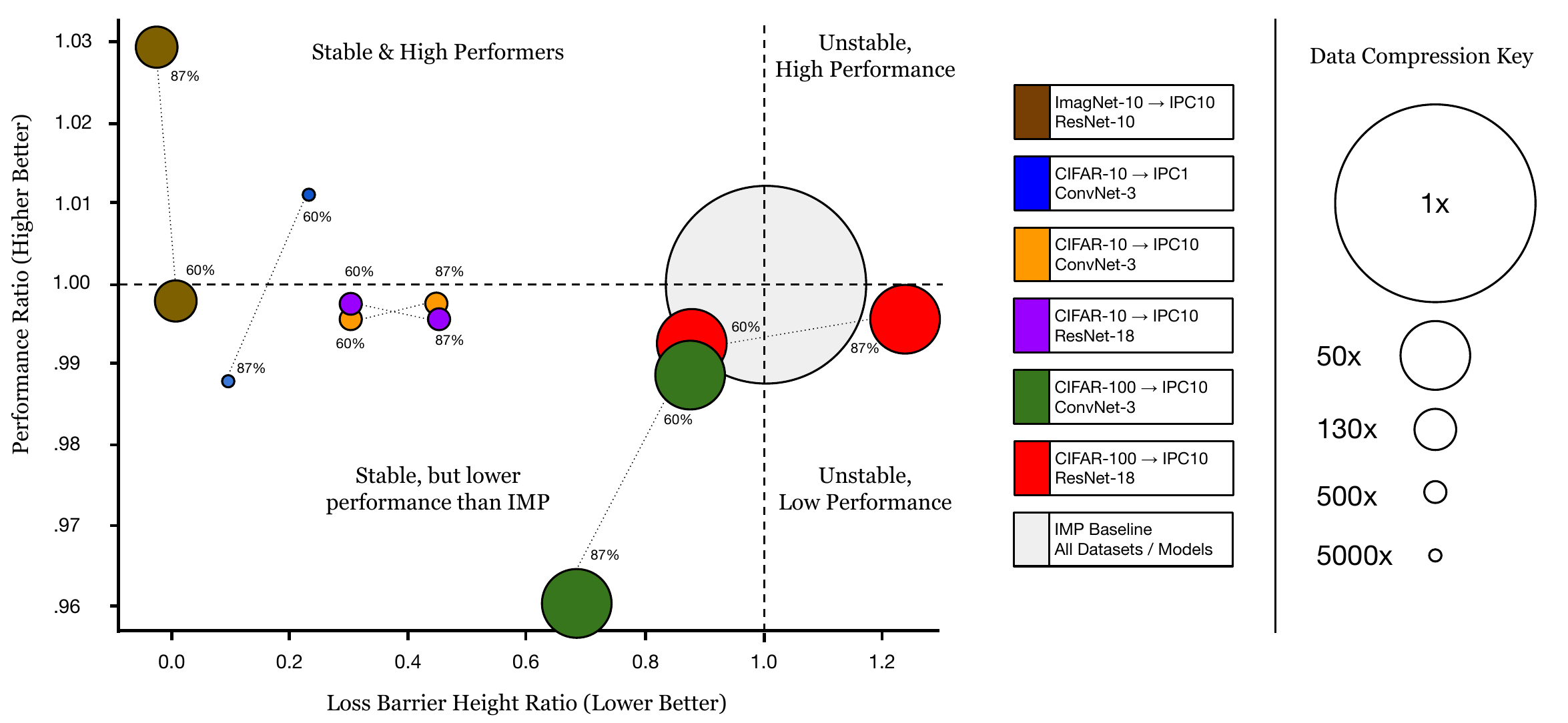}}
  \caption{Comparison of synthetic subnetworks and IMP subnetworks across models \& datasets. Both the 60\% and 87\% sparsity model is chosen for each example and connected with a dotted line. These are compared to IMP's 60\% and 87\% sparsity respectively. The size of the marker represents the amount of data compression through dataset distillation used to find the sparsity mask. For example, distilling CIFAR-10 (default 5,000 images per class) down to 10 images per class is a 500x compression ratio. Performance is measured as the ratio of test accuracy between Synthetic and IMP subnetworks. Loss Barrier Height is measured by subtracting the trained loss and the halfway interpolated trained loss from linear interpolation experiments. }
  \label{fig:scatterplot}
\end{figure}

For a given dataset, as compression ratio increases, we see more stability than IMP; however, the performance trend largely depends on the distilled accuracy in \citet{EfficientDatasetCondensation}. As the complexity of the dataset increases, dataset distillation under-performs. Most notably, we achieve full linear mode connectivity for ConvNet-3 on CIFAR-10 and ResNet-10 on Imagenet-10. Although multiple factors influence stability, \citet{EfficientDatasetCondensation} optimized the synthetic data specifically for these models, suggesting that high-quality synthetic data enhances stability. As distillation methods improve, we expect to preserve more label information (pushing the points up) and be able to use fewer images per class (pushing the points to the left). Until these methods improve, distilled pruning cannot yet compete as a standalone method for performance, only for finding stable subnetworks.

\section{Stable Initializations \& The Lottery Ticket Hypothesis}
Distilled Pruning and traditional IMP can be used in tandem to leverage their respective strengths. Traditional IMP fails to find high-performing subnetworks when the dense initialization is unstable. During the iterative retraining process, the optimizer cannot navigate to the same minima, implying that the sparsity mask generated from the previous pruning process is ill-informed for this area of the loss landscape. To address this shortcoming, we find a stable subnetwork from the unstable dense model using distilled pruning. Then, we apply traditional IMP to this stable, sparse initialization.

\begin{figure}[htb]
  \centering
  \makebox[0pt]{\includegraphics[width=2.75in]{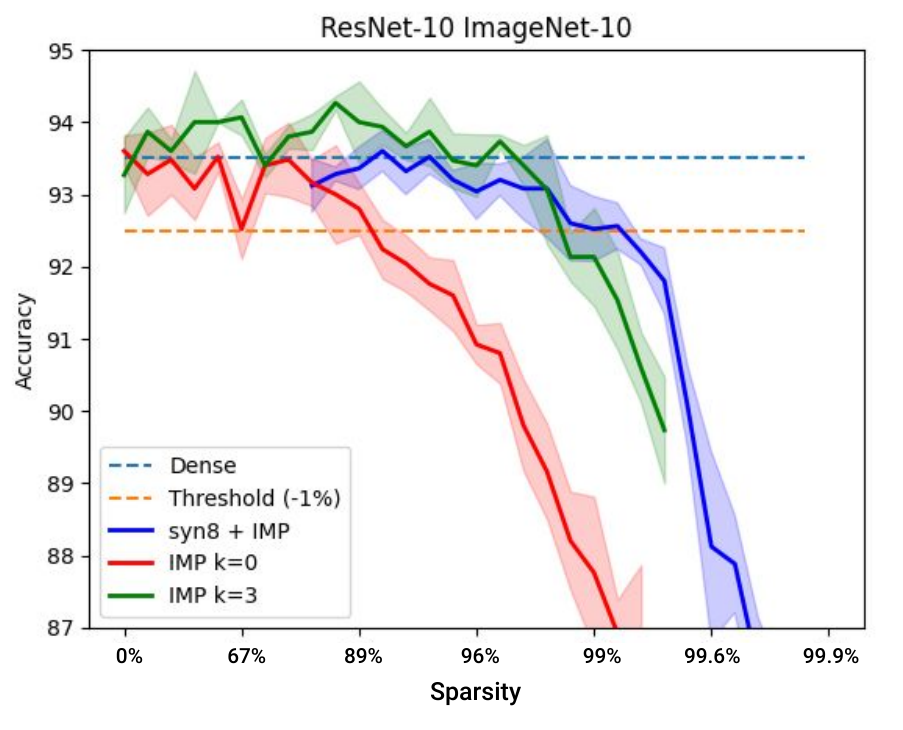}}
  \caption{Performance comparison of traditional IMP ($k=0$), IMP with weight rewinding to the 3rd epoch in training ($k=3$), and traditional IMP performed on a stable, synthetic subnetwork at 83\% sparsity. We emphasize IMP $k=3$ is not a lottery ticket as we are not training from initialization. Each run uses a comparable computation cost as the 8 iterations of distilled pruning is negligible compared to a single training run.}
  \label{fig:syn+imp}
\end{figure}

In Figure \ref{fig:syn+imp}, Traditional IMP (IMP k=0 in red) cannot find a lottery ticket past $90\%$ sparsity. To generate a stable sparsity mask, we prune our dense model for 8 steps with distilled pruning, denoted as syn8. Then, we apply IMP to this subnetwork (in blue), still rewinding the weights back to initialization. Using both methods together finds true lottery tickets at sparsities higher than the vanilla lottery ticket hypothesis addresses.

To compare our method of stabilizing the initialization, we also measure the performance of IMP with rewinding to an early point in training \citep{LMCLTH}, denoted as IMP k=3 for spawning 3 epochs in. We emphasize that rewinding to early training is \textbf{not a lottery ticket}; lottery tickets must be sparsely trained from initialization. We show that our lottery tickets observe similar performance at high sparsities when compared to the matching subnetworks produced in spawning.

\section{Discussion}
The synthetic sparsity masks are a result of pruning irrelevant parameters on an already compressed dataset. We believe the weights that appear in the IMP subnetworks, but not synthetic subnetworks, contribute to sharper landscapes. We validate this across linear mode connectivity studies, visualizations of the loss landscape, and a Hessian analysis. Distilled pruning has shown to increase stability; however, this often can lead to worse performance in cases where the distilled data is not aligned well to the training set such as CIFAR-100.  

Analyzing these sparsity masks in isolation did not reveal clear structural patterns. Distilled Pruning finds masks to have extremely similar sparsity per layer with a seemingly independent distribution of pruned parameters when compared to IMP. We invite future researchers to engage in this analysis. 

This work is an initial step into exploring the impact of using synthetic data, specifically distilled data, to improve pruning. Since this work shows how the data used to prune directly impacts the loss landscape and thus the generalizability, we aim to rethink the use of any data in these scenarios, not just through dataset distillation. For example, pruning models by retraining with knowledge distillation \cite{nemotron} may observe similar properties to this work. 

Lastly, we reopen the idea of finding lottery tickets at initialization on larger models by focusing on pruning parameters leading to instability first. To the best of our knowledge, this work is the largest case of stability at initialization without weight permutation. The Lottery Ticket Hypothesis \citep{LTH} has only been validated up to the limits of traditional IMP. We expand upon these limits, showing that higher sparsities are indeed possible through the combination of distilled pruning and IMP. Due to the under performance of dataset distillation, we believe that a new algorithm, perhaps using knowledge distillation from the dense model, will be essential for validating the lottery ticket hypothesis on large, real-world settings. We hope our contribution of finding stable subnetworks in unstable dense models provides the stepping stone for such advancements.

\bibliography{iclr2025_conference}
\bibliographystyle{plainnat}
\newpage
\appendix

\end{document}

%% file: math_commands.tex
\usepackage{amsmath,amsfonts,bm}

\def\eqref#1{equation~\ref{#1}}

\def\1{\bm{1}}

\DeclareMathAlphabet{\mathsfit}{\encodingdefault}{\sfdefault}{m}{sl}
\SetMathAlphabet{\mathsfit}{bold}{\encodingdefault}{\sfdefault}{bx}{n}

\DeclareMathOperator*{\argmin}{arg\,min}